# Estimation and Clustering with Infinite Rankings

**Marina Meilă** and **Le Bao**
Department of Statistics
University of Washington
Seattle, WA 98195-4322
{mmp,lebao}@stat.washington.edu

## Abstract

This paper presents a natural extension of stagewise ranking to the the case of infinitely many items. We introduce the infinite generalized Mallows model (IGM), describe its properties and give procedures to estimate it from data. For estimation of multimodal distributions we introduce the *Exponential-Blurring-Mean-Shift* nonparametric clustering algorithm. The experiments highlight the properties of the new model and demonstrate that infinite models can be simple, elegant and practical.

## 1 Introduction

The stagewise ranking model of [6], also known as *generalized Mallows (GM)*, has been recognized as particularly appropriate for modeling the human process of ranking. This model assigns a permutation $\pi$ over $n$ items a probability that decays exponentially with its distance to a *central permutation* $\sigma$. Here we study this class of models in the limit $n \to \infty$, with the assumption that out of the infinitely many items ordered, one only observes those occupying the first $t$ ranks.

Ordering an infinite number of items is common in retrieval tasks: search engines, programs that match a face, or a fingerprint, or a biological sequence against a database, all output the first $t$ items in a ordering over virtually infinitely many objects. We shall call this output a *top-t* ordering. Unlike machines, people can only reliably rank a small number of items. The GM model has been successfully used to model human ranking decisions. We can view the difference between the standard GM model and the *infinite GM model* that we introduce here as the difference between an election where each voter returns an ordering of a small number of preselected candidates (nominees) and a "grassroots" election process, where everyone can nominate and order their own favourites from a virtually unlimited population. For instance, the difference between "Order the following issues by how much you care about them" vs. "List in order the issues that you care most about" illustrates the difference between the standard and the infinite GM models. By these examples, we argue that the infinite GM corresponds to realistic scenarios.

After defining the infinite GM model, we show that it has sufficient statistics and give algorithms for estimating its parameters from data in the Maximum Likelihood (ML) framework. To be noted that our model will have an infinite number of parameters, of which only a finite number will be constrained by the data from any finite sample. The existence of sufficient statistics also enables us to construct and characterize the conjugate prior for this class of models.

Then, we consider the clustering of top-$t$ ranking data. and introduce an adapted version of the well known Gaussian Blurring Mean-Shift algorithm [2] (GBMS) that we call Exponential Blurring Mean Shift (EBMS).

## 2 Finite and infinite permutations and their top-$t$ orderings

We consider permutations $\sigma$ over the set of positive natural numbers $\mathbb{N}^* = \{1, 2, \ldots, i \ldots\}$. Following standard notation, $\sigma(i)$ denotes the rank of item $i$, $\sigma^{-1}(j)$ the item at rank $j$ in $\sigma$. The permutation matrix $\Sigma$ corresponding to $\sigma$ has $\Sigma_{ij} = 1$ iff $\sigma(i) = j$. For any two permutations $\sigma, \sigma'$ over $\mathbb{N}^*$, the matrix product $\Sigma\Sigma'$ corresponds to the function composition of permutations $\sigma'(\sigma)$. In our case, $\Sigma, \Sigma'$ will be infinite matrices with exactly one 1 in every row and column.

A top-$t$ ordering $\pi^{-1}$ is the prefix $(\pi^{-1}(1) \ldots \pi^{-1}(t))$ of some infinite ordering. The notation $\pi^{-1}$ indicates that we observe items, not ranks. For the rest of this paper, the term *ordering* will denote the inverse of a permutation, i.e. the list of items associated to ranks

1, 2, .... In general, a Greek letter $\pi$ or $\sigma$ denotes a permutation, also called *ranking*, while the symbols $\pi^{-1}, \sigma^{-1}$ denote the corresponding orderings. Further, our notation distinguishes when possible between observed orderings, denoted by $\pi^{-1}$, which by virtue of being observed, are always truncated, and the "central permutations", ideal infinite objects denoted by $\sigma$. What we try to estimate is a top-$t'$ ordering of $\sigma$ and this is denoted by $\sigma^{-1}$.

The matrix $\Pi$ of a top-$t$ ordering $\pi$ has $t$ columns, each with an infinite number of zeros and with a 1 in row $\pi^{-1}(j), j = 1 : t$. Rigourously this matrix should be denoted $\Pi^T$ since it is the matrix of $\pi^{-1}$ but we opt for $\Pi$ to simplify notation. For a permutation $\sigma$ and a top-$t$ ordering $\pi^{-1}$, the matrix $\Pi^T \Sigma$ corresponds to the composition $\sigma(\pi^{-1})$ listing the ranks in $\sigma$ of the items in $\pi^{-1}$. We shall use its transpose, $\Sigma^T \Pi$ which is the $\infty \times t$ matrix formed with rows $\pi^{-1}(1) \ldots \pi^{-1}(j)$ of $\Sigma$ as columns. For any $\Sigma^T \Pi$ matrix, its *code* $(s_j, j = 1 : t)$ is defined as follows: $1 + s_j$ is the rank of $\pi^{-1}(j)$ in $\sigma_{|\mathbb{N}^* \setminus \{\pi^{-1}(1), \ldots \pi^{-1}(j-1)\}}$. Thus, $s_1$ is the number of 0's preceding the 1 in the first column of $\Sigma^T \Pi$; after we delete the row containing this 1, $s_2$ is the number of 0's preceding the 1 in the second column; after deleting the row containing this 1, $s_3$ is the number of 0's preceding the 1 in the third column, and so on. Hence $s_j \in \{0, 1, 2, \ldots\}$,

$$s_j(\Sigma^T \Pi) = \sigma(\pi^{-1}(j)) - 1 - \sum_{j' < j} 1_{[\sigma(\pi^{-1}(j')) < \sigma(\pi^{-1}(j))]}$$

We now introduce the distance

$$d_\theta(\pi^{-1}, \sigma) = \sum_{j=1}^{t} \theta_j s_j(\Sigma^T \Pi) \quad (1)$$

with $\theta = (\theta_{1:t})$ a vector of *strictly positive* parameters; $d_\theta$ is the extension to infinite orderings of the $d_\theta$ of [6]. In general, this "distance" between a top-$t$ ordering and an infinite ordering is not a metric. When $\theta_j$ are all equal $d_\theta(\pi^{-1}, \sigma) = \theta d_K(\pi^{-1}, \sigma)$, with the $d_K$ known as the Kendall distance. $d_K$ counts the number of adjacent transpositions needed to make $\sigma$ compatible with $\pi^{-1}$ and is a metric.

## 3 The Infinite Generalized Mallows model

Now we are ready to introduce the *infinite generalized Mallows (IGM)* model. We start with the observation that as any top-$t$ ordering can be represented uniquely by a sequence of $t$ natural numbers, defining a distribution over the former is equivalent to defining a distribution over the latter, which is a more intuitive task. In keeping with the GM paradigm, we shall choose $t$-sequences for which each $s_j$ is sampled independently from a discrete exponential with parameter $\theta_j > 0$.

$$P(s_j) = \frac{1}{\psi(\theta_j)} e^{-\theta_j s_j}, \quad s_j = 0, 1, 2, \ldots \quad (2)$$

The normalization constant is $\psi(\theta_j) = \sum_{k=0}^{\infty} e^{-\theta_j k} = \frac{1}{1 - e^{-\theta_j}}$. The IGM model can now be defined as

$$P_{\theta, \sigma}(\pi^{-1}) = e^{-\sum_{j=1}^{t} [\theta_j s_j (\Sigma^T \Pi) + \ln \psi(\theta_j)]} \quad (3)$$

The distribution $P_{\theta, \sigma}$ has a $t$-dimensional real parameter $\theta$ and an infinite-dimensional discrete parameter $\sigma$. Any top-$t$ ordering $\pi^{-1}$ stands for a set of infinite sequences starting with $s_{1:t}$, $P_{\theta, \sigma}$ can be viewed as the marginal of $s_{1:t}$ in the infinite product space defined by the distribution $P_{\theta, \sigma}(s) = e^{-\sum_{j=1}^{\infty} [\theta_j s_j + \ln \psi(\theta_j)]}, s \in \mathbb{N} \times \mathbb{N} \times \ldots$. In contrast with the finite GM, the parameters $\theta_j$ must be strictly positive in order for the probability distribution to exist. The most probable $\pi^{-1}$ for any given $t$ is the top-$t$ ordering which corresponds to $s_1 = \ldots = s_t = 0$. This is the top-$t$ prefix of $\sigma^{-1}$.

The ordering $\sigma$ is called the *central permutation* of $P_{\theta, \sigma}$. The parameters $\theta$ control the spread around the mode $\sigma$. Larger $\theta_j$ correspond to more concentrated distributions. These facts are direct extensions of statements about the GM model from [6] and therefore the detailed proofs are omitted.

## 4 Estimating the model from data

We are given a set of $N$ top-$t$ orderings $\mathcal{D}$. Each $\pi^{-1} \in \mathcal{D}$ can have a different length $t$; all $\pi^{-1}$ are sampled independently from a $P_{\theta, \sigma}$ with unknown parameters. We propose to estimate $\theta, \sigma$ from this data in the ML paradigm. We will start by rewriting the log-likelihood of the model, in a way that will uncover a set of sufficient statistics. Then we will show how to estimate the model based on the sufficient statistics.

### 4.1 Sufficient statistics

The following result lets us understand the structure of the log-likelihood and is thus key to the discrete optimization over $\sigma$.

For any square (infinite) matrix $A \in \mathbb{R}^{\mathbb{N}^* \times \mathbb{N}^*}$, denote by $L(A)$ the sum of the elements below the diagonal of $A$, i.e in the *lower triangle* of $A$. Let $L_\sigma(A) = L(\Sigma^T A \Sigma)$, and let $\mathbf{1} \in \mathbb{R}^{\mathbb{N}^*}$ be the vector of all 1's. For any $\pi^{-1}$ let $t_\pi$ be its length and denote $t_{max} = \max_\mathcal{D} t_\pi, T = \sum_\mathcal{D} t_\pi$.

**Theorem 1**

$$\ln P_{\theta,\sigma}(\mathcal{D}) = -\sum_{j\geq 1}[\theta_j L_\sigma(R_j) + N_j \ln \psi(\theta_j)] \quad (4)$$

where $R_j = q_j \mathbf{1}^T - Q_j$, and $N_j$ is the number of $\pi^{-1} \in \mathcal{D}$ that have length $t_\pi \geq j$ (in other words, that contain rank $j$); $q_j = [q_{i,j}]_{i \in \mathbb{N}^*}$, with $q_{i,j}$ being the number of times $i$ is observed in rank $j$ in the data $\mathcal{D}$, $Q_j = [Q_{ii',j}]_{i',i \in \mathbb{N}^*}$ is a matrix whose element $Q_{ii',j}$ counts how many times $\pi(i) = j$ and $\pi(i') < j$.

**Proof** Let $Q_0$ be the infinite matrix that has 1 above the main diagonal and 0 elsewhere, $(Q_0)_{ij} = 1$ iff $j > i$ and let $\Pi_{:j}$ denote the $j$-th column of $\Pi$. By definition, $s_j$ represents the number of 0's preceding 1 in column $j$, minus all the 1's in the submatrix $\Sigma^T \Pi_{1:\sigma(\pi^{-1}(j)),1:j}$, i.e $s_j(\Sigma^T \Pi) = \sum_{l \geq 1}(Q_0 \Sigma^T \Pi_{:j})_l (\mathbf{1} - \Sigma^T \Pi_{:1} - \Sigma^T \Pi_{:2} - \ldots \Sigma^T \Pi_{:j-1})_l = (\mathbf{1} - \sum_{j'<j} \Sigma^T \Pi_{:j'})^T Q_0 \Sigma^T \Pi_{:j}$
$= \mathbf{1}^T Q_0 \Sigma^T \Pi_{:j} - \sum_{j'<j} \Pi_{:j'}^T \Sigma Q_0 \Sigma^T \Pi_{:j}$
$= \text{trace } Q_0 \Sigma^T [\Pi_{:j} \mathbf{1}^T - \sum_{j'<j} \Pi_{:j} \Pi_{:j'}^T \Sigma]$
$= L(\Sigma^T [\Pi_{:j} \mathbf{1}^T - \sum_{j'<j} \Pi_{:j} \Pi_{:j'}^T] \Sigma) = L_\sigma(\Pi_{:j} \mathbf{1}^T - \sum_{j'<j} \Pi_{:j} \Pi_{:j'}^T)$ In the first equality we use the fact that multiplying left by $Q_0$ counts the zeros preceding 1 in a column, and in the last we use the identity $\mathbf{1}^T \Sigma = \mathbf{1}^T$. Because $L_\sigma$ is a linear operator, summing over $\pi^{-1} \in \mathcal{D}$ yields $\sum_{\pi^{-1} \in \mathcal{D}} s_j(\Sigma^T \Pi) = L_\sigma([\sum_{\pi^{-1} \in \mathcal{D}} \Pi_{:j}] \mathbf{1}^T - [\sum_{\pi^{-1} \in \mathcal{D}} \sum_{j'<j} \Pi_{:j} \Pi_{:j'}^T])$. It is easy to verify now that the first sum represents $q_j$ and the second one represents $Q_j$. By setting $R_j(\pi^{-1}) = Q_j(\pi^{-1}) = 0$ for $j > t_{max}$, we can write the log-likelihood as in (4). □

**Corollary 2** If $\theta_1 = \theta_2 = \ldots = \theta$ then the log-likelihood of the data $\mathcal{D}$ can be written as

$$\ln P_{\theta,\sigma}(\mathcal{D}) = -\theta L_\sigma(R) - T \ln \psi(\theta) \quad (5)$$

with $R = q\mathbf{1}^T - Q$ and $q = \sum_j q_j$, $Q = \sum_j Q_j$.

Note that $q_i, Q_{ii'}$ represent respectively the number of times item $i$ is observed in the data and the number of times item $i'$ precedes $i$ in the data.

Theorem 1 and its corollary 2 show that the infinite model $P_{\theta,\sigma}$ has *sufficient statistics*. As $\theta, \sigma$ of the model are infinite, the sufficient statistics $R_j, N_j$ (or $R$) are infinite too. However, for any finite data set, these matrices and vector will contain only a finite number of non-zero entries. Another consequence is that the data will only constrain a finite number of parameters of the model. The log-likelihood (4) depends only on the parameters $\theta_{1:t_{max}}$. Maximizing likelihood will determine $\theta_{1:t_{max}}$ leaving the other $\theta_j$ parameters undetermined.

Let $n$ be the number of distinct items observed in the data. From $\sigma$, we can estimate at most its restriction to the items observed, i.e. the restriction of $\sigma$ to the set $\bigcup_\mathcal{D} \pi^{-1}$. In other words the data determine a top-$n$ (partial) ordering corresponding to $\sigma^{ML}$, leaving the rest undetermined[1].

Note also that, unlike in the finite case, the sufficient statistics do not necessarily compress the data. They can take up more space than storing the permutations, and their size grows with $N$.

### 4.2 ML estimation: the case of a single $\theta$

We now go on to the practical estimation of $\theta$ and $\sigma$ starting with the case of equal $\theta_j$, i.e $\theta_1 = \theta_2 = \ldots = \theta$. In this case, equation (5) shows that the estimation of $\theta$ and $\sigma$ decouple. For any fixed $\sigma$, equation (5) attains its minimum at

$$\theta = \ln(1 + T/L_\sigma(R)). \quad (6)$$

In contrast to the above simple formula, for the finite GM, the likelihood has no analytic solution for $\theta$ [6]. The estimated value of $\theta$ increases when $L_\sigma(R)$ decreases. In other words, if the lower triangle of $\Sigma^T R \Sigma$, counting the "out of order" pairs, has very low counts, we conclude that the distribution is very concentrated, hence has a high $\theta$.

Estimating $\sigma^{ML}$ amounts to minimizing $L_\sigma(R)$ w.r.t $\sigma$, independently of the value of $\theta$. The optimal $\sigma$ according to Corollary 2 is the (partial) permutation that minimizes the lower triangular part of $\Sigma^T R \Sigma$[2]. To find it we exploit an idea first introduced in [11]. This idea is to search for $\sigma^{-1} = (i_1, i_2, i_3, \ldots)$ in a stepwise fashion, starting from the top item $i_1$ and continuing down.

Assuming for a moment that $\sigma^{-1} = (i_1, i_2, i_3, \ldots)$ is known, the cost to be minimized $L_\sigma(R)$ can be decomposed columnwise as $L_\sigma(R) = \sum_{l \neq i_1} R_{li_1} + \sum_{l \neq i_1,i_2} R_{li_2} + \sum_{l \neq i_1,i_2,i_3} R_{li_3} + \ldots$ where the number of non-trivial terms is one less than the dimension of $R$. It is on this decomposition that the search algorithm is based. If $\sigma^{-1}$ is not known, a search algorithm could try every $i_1$ in turn, saving the partial sums, then for a chosen $i_1$ value could try all $i_2$'s that could follow it, etc. This type of search is represented by a *search tree*, whose nodes are candidate prefixes for $\sigma^{-1}$.

---

[1] That not even the restriction to the observed items is always completely determined can be seen by the following example. Assume the data consists of the the two top-$t$ orderings $(a, b, c)$, $(a, b, d)$. Then $(a, b, c, d)$ and $(a, b, d, c)$ are both ML estimates for $\sigma^{-1}$; hence, it would be more accurate to say that the ML estimate of $\sigma^{-1}$ is the *partial ordering* $(a, b, \{c, d\})$.

[2] If the optimum is a partial permutation $\hat{\sigma}^{-1}$ then any permutation compatible with $\hat{\sigma}^{-1}$ will be a minimizer of $L_\sigma(R)$.

For our problem, the search tree has $n!$ terminal nodes, one for each possible ordering of the observed items. Finding the lowest cost path through the tree is equivalent to minimizing $L_\sigma(R)$. *Branch-and-bound (BB)* [12] algorithms are methods to explore the tree nodes in a way that guarantees that the optimum is found, even though the algorithm may not visit all the nodes in the tree. The number of nodes explored in the search for $\sigma^{-1}$ depends on the sufficient statistics matrix $R$. In the worst case, the number of nodes searched can be a significant fraction of $n!$ and as such intractable for all but small $n$. However, if the data distribution is concentrated around a mode, then the search becomes tractable.

We call the BB algorithm for estimating $\sigma$ the BBoundR. The algorithm's main steps, characteristics of a BB algorithm, are given in figure 1. In addition to the exact BBoundR algorithm, various heuristic search techniques can be used. Two of them which showed good performance for the standard GM model are the greedy (depth-first) search and the heuristic of [7]. In the latter, one obtains $\sigma$ by sorting $r_l = \sum_k R_{kl}$ in increasing order[3].

In conclusion, to estimate the parameters from data in a single parameter case, one first computes the sufficient statistics, then a prefix of $\sigma^{-1}$ is estimated by BBoundR or heuristic methods, and finally, with the obtained ordering of the observed items, one can compute the estimate of $\theta$.

### 4.3 ML estimation: the case of general $\theta$

Maximizing the likelihood of the data $\mathcal{D}$ is equivalent, by Theorem 1, with minimizing

$$J(\theta, \sigma) = \sum_j [\theta_j L_\sigma(R_j) + N_j \ln \psi(\theta_j)] \qquad (7)$$

This estimation equation does not decouple w.r.t $\theta$ and $\sigma$. Minimization is however possible, due to the following two observations. First, for any fixed set of $\theta_j$ values, minimization w.r.t $\sigma$ is possible by the algorithms described in the previous section. Second, for fixed $\sigma$, the optimal $\theta_j$ parameters can be found analytically by $\theta_j = \ln(1 + N_j/L_\sigma(R_j))$.

The two observations immediately suggest an alternate minimization approach to minimizing $J$. The algorithm is given in Figure 2. As both steps increase the likelihood, the algorithm will stop in a finite number of steps at a local optimum

---

[3]Described here is a simplified version of the SortRowsR heuristic. The algorithm as proposed by [7] also performs limited search around the obtained permutation.

---

**Algorithm** BBoundR

**Input** Matrix $R$

The algorithm maintains a priority queue $Q$ storing search tree nodes. For each node $\rho$, we store: the path to the node $(r_{1:j})$, the cost $C$ of this path, a lower bound $A$ on the cost-to-go, and the sum $T = C + A$. Nodes are prioritized by $T$.

While $Q$ is not empty:

1. $\rho \leftarrow$ Extract $\min(Q)$
2. if $j(\rho) = n - 1$ **Output** $\rho$. **Stop**
3. else, for $k = 1 : n - j$
   (a) Create child $\rho'$ of $\rho$
   (b) Evaluate $C(\rho'), A(\rho'), T(\rho')$
   (c) enqueue $\rho'$ in $Q$

Figure 1: Algorithm BBoundR.

---

**Algorithm** EstimateSigmaTheta

**Input** Sufficient statistics $R_j, N_j$, $j = 1 : t_{max}$
Initial parameter values $\theta_{1:t_{max}} > 0$

1. Iterate until convergence:
   (a) Calculate $R_\theta = \sum_j \theta_j R_j$
   (b) Find partial ordering $\sigma^{-1} = \mathrm{argmin}_\sigma L_\sigma(R_\theta)$ by BBoundR
   (c) Estimate $\theta_j = \ln[1 + N_j/L_\sigma(R_j)]$ for $j = 1 : t_{max}$

**Output** $\sigma^{-1}$, $\theta_{1:t_{max}}$

Figure 2: Algorithm EstimateSigmaTheta.

## 5 Clustering

Having defined a distance and a method for estimating ML parameters gives one access to a large number of the existing clustering paradigms originally defined for Euclidean data. For instance, the extensions of the K-means and EM algorithms to infinite orderings is immediate, and so are extensions to other distance-based clustering methods. In addition, we introduce a nonparametric clustering method, the *Exponential Blurring Mean-Shift (EBMS)*. Nonparametric clustering is motivated by the fact that in many real applications the number of clusters is unknown and outliers exist. We adapt for ranked data the well known blurring mean-shift algorithm [2]. We choose the exponential kernel $K_\theta(\pi, \sigma) = e^{-\theta d_K(\pi, \sigma)}$. For a data set of top-$t$ rankings, the Nadaraya-Watson kernel estimator is $\widehat{r}(\pi_i) = \sum_{j=1}^{n} \frac{\exp^{-\theta d(\pi_i, \pi_j)}}{\sum_{i=1}^{n} \exp^{-\theta d(\pi_i, \pi_j)}} \pi_j$.

Table 1: Results of estimation experiments. Top: mean and standard deviation of $\theta^{ML}$ for two values of the true $\theta$ and for different $t$ values and sample sizes $n$. Bottom: the proportion of cases when the ordering error, i.e the number inversions w.r.t the true $\sigma^{-1}$ was 0, respectively 1. Each estimation was replicated 25 or more times.

| Estimates of $\theta$ (mean stdev) | | | | | | | |
|---|---|---|---|---|---|---|---|
| $\theta$ | N | 200 | | 500 | | 2000 | |
| | | mean | std | mean | std | mean | std |
| 0.69 | $t=2$ | 0.68 | 0.04 | 0.68 | 0.03 | 0.68 | 0.024 |
| | $t=4$ | 0.67 | 0.03 | 0.69 | 0.02 | 0.69 | 0.01 |
| | $t=8$ | 0.68 | 0.02 | 0.69 | 0.01 | 0.69 | 0.007 |
| 1.38 | $t=2$ | 1.34 | 0.13 | 1.37 | 0.09 | 1.37 | 0.04 |
| | $t=4$ | 1.40 | 0.06 | 1.38 | 0.05 | 1.38 | 0.03 |
| | $t=8$ | 1.37 | 0.03 | 1.38 | 0.03 | 1.38 | 0.01 |

| Ordering error | | | | | | | |
|---|---|---|---|---|---|---|---|
| $\theta$ | N | 200 | | 500 | | 2000 | |
| | | $d_K=0$ | $d_K=1$ | $d_K=0$ | $d_K=1$ | $d_K=0$ | $d_K=1$ |
| 0.69 | $t=2$ | 0.36 | 0.38 | 0.32 | 0.38 | 0.24 | 0.36 |
| | $t=4$ | 0.28 | 0.46 | 0.36 | 0.28 | 0.32 | 0.42 |
| | $t=8$ | 0.40 | 0.38 | 0.36 | 0.34 | 0.40 | 0.30 |
| 1.38 | $t=2$ | 0.78 | 0.18 | 0.76 | 0.18 | 0.78 | 0.18 |
| | $t=4$ | 0.80 | 0.20 | 0.82 | 0.14 | 0.80 | 0.16 |
| | $t=8$ | 0.84 | 0.14 | 0.72 | 0.22 | 0.92 | 0.08 |

The EBMS algorithm is summarized in Figure 3. It shift the "points" (i.e top-$t$ orderings) to new locations obtained by a locally weighted combination of all the data. Thus, every $\pi^{-1}$ is "attracted" towards its closest neighbors; as the shifting is iterated the data collapse into one or more clusters. The *scale parameter* $\theta$ influences the size of the local neighborhood of a top-$t$ ordering, and thereby controls the granularity of the final clustering: for small $\theta$ values (large neighborhoods), points will coalesce more and few large clusters will form; for large $\theta$'s the orderings will cluster into small clusters and singletons. In the EBMS algorithm, we estimate the scale parameter $\theta$ at each iteration by solving the equation in step (d). The l.h.s of this equation represents the average distance in the data set, while the r.h.s is the expected distance to the centroid under the Infinite GM model for permutations of length $t_\pi$.

In step 5c of the algorithm, the new ranking can be much longer than the original partial ranking. As will be shown in section 7, the last ranks are subject to noise and overfitting. Therefore we truncate the new ranking back to $t$.

Since at each step we round $\widehat{r}(\pi_i)$ to the closest permutation, the algorithm will stop in a finite number of steps, when no ordering moves from its current position.

In the algorithm one evaluates distances between top-

**Algorithm** EBMS

**Input** Top-$t$ orderings $\mathcal{D} = \{\pi_i^{-1}\}_{i=1:N}$ with same length $t_\pi$.

1. Count the distinct permutations to obtain reduced $\tilde{\mathcal{D}}$ and counts $n_i \geq 1$ for each ordering $\pi_i^{-1} \in \tilde{\mathcal{D}}$.
2. For $\pi_i^{-1} \in \tilde{\mathcal{D}}$ compute $q_i, Q_i, R_i$ the sufficient statistics of a single data point.
3. For $\pi_i^{-1}, \pi_j^{-1} \in \tilde{\mathcal{D}}$ calculate Kendall distance $d_{ij} = d_K(\pi_i^{-1}, \pi_j^{-1})$
4. Set the scale $\theta$ by solving the equation

$$\frac{2}{\tilde{N}(\tilde{N}-1)} \sum_{i<j} d_{ij} = \frac{t_\pi \times e^{-\theta}}{1 - e^{-\theta}} - \sum_{j=1}^{t_\pi} \frac{j \times e^{-\theta}}{1 - e^{-j\theta}}$$

5. For $\pi_i \in \tilde{\mathcal{D}}$     *(Compute weights and shift)*

   (a) For $\pi_j \in \tilde{\mathcal{D}}$: set $\alpha_{ij} = \frac{\exp(-\theta d_{ij})}{\sum_{j'=1}^{n} \exp(-\theta d_{ij'})}$
   (b) Calculate $\bar{R}_i = \sum_{\pi_j \in \tilde{\mathcal{D}}} n_j \alpha_{ij} R_j$
   (c) Estimate $\sigma_i^{-1}$ the "central" permutation that optimizes $\bar{R}_i$ by BBOUNDR or by heuristics
   (d) Set $\pi_i^{-1} \leftarrow \sigma_i^{-1}(1:t_\pi)$

6. Repeat from step 1 until no $\pi_i^{-1}$ changes.

**Output** $\tilde{\mathcal{D}}$

Figure 3: The EBMS algorithm.

$t$ orderings. We define the distance $d(\pi, \pi')$ to be the distance between the sets of infinite orderings compatible with $\pi^{-1}$ respectively $(\pi')^{-1}$. We need to extend the Kendall distance to cover this case and for this we adopt an idea from [4] which lets us express and evaluate $d(\pi, \pi')$ efficiently. The details were omitted for lack of space but they are available in the full paper [10].

## 6 The conjugate prior

The existence of sufficient statistics implies the existence of a conjugate prior for the parameters of model (3) [5]. Here we introduce the general form of this prior and show that computing with the conjugate prior (or posterior), is significantly harder than computing with the likelihood (3).

We shall assume for simplicity that all top-$t$ rankings have the same $t$. Consequently, our parameter space consists of the real positive vector $\theta_{1:t}$ and the discrete infinite parameter $\Sigma$. Let $\nu$ denote the *prior strength*, representing the equivalent sample size, and

$\lambda_1, \Lambda_j, j = 2 : t$ be the prior parameters corresponding to the sufficient statistics $q_1, Q_{2:t}$, normalized.

**Proposition 3** *Let $\nu > 0$, $\lambda_1$ be a vector and $\Lambda_j, j = 2 : t$ denote a set of possibly infinite matrices satisfying $\lambda_1 \geq 0$; $\Lambda_{ii',j} \geq 0$, for all $i, i', j$; $\mathbf{1}^T \Lambda_j \mathbf{1} = (j - 1)$ for all $j > 1$. Denote $\mathbf{\Lambda} = \{\nu, \lambda_1, \Lambda_{2:t}\}$ and $R_j^0 = \Lambda_j \left(\frac{1}{j-1}\mathbf{1}\mathbf{1}^T - I\right)$ for $j > 1$, $R_1^0 = \lambda_1 \mathbf{1}^T$. Then, the distribution*

$$P_{\mathbf{\Lambda}}(\sigma, \theta) \propto e^{-\nu \sum_{j=1}^t [\theta_j L(\Sigma^T R_j^0 \Sigma) + \ln \psi(\theta_j)]} \quad (8)$$

*is a conjugate prior for the model $P_{\theta,\sigma}(\pi^{-1})$ in (3).*

**Proof** Given observed permutations $(\pi^{-1})_{1:N}$, the posterior distribution of $(\sigma, \theta)$ is updated by $P(\theta, \sigma \,|\, \mathbf{\Lambda}, (\pi^{-1})_{1:N}) \propto \exp\left(-\sum_{j=1}^t [(\nu L_\sigma(R_j^0) + L_\sigma(R_j))\theta_j + (N+\nu)\ln\psi(\theta_j)]\right)$ $\exp\left(-(N+\nu)\sum_{j=1}^t [\theta_j L_\sigma\left(\frac{\nu R_j^0 + R_j}{N+\nu}\right) + \ln\psi(\theta_j)]\right)$. If the hyperparameters $\nu, \lambda_1, \Lambda_{2:t}$ satisfy the conditions of the proposition, then the new hyperparameters $\mathbf{\Lambda}' = \{\nu + N, (\nu\lambda_1 + q_1)/(\nu + N), (\nu\Lambda_j + R_j)/(\nu + N), j = 2 : t\}$ satisfy the same conditions. $\square$

The conjugate prior is defined in (8) only up to a normalization constant. This normalization constant is not always computable in closed form. Another aspect of conjugacy is that one prefers the conjugate hyperparameters to represent expectations of the sufficient statistics under some $P_{\theta,\sigma}$. The conditions in Proposition 3 are necessary, but not sufficient to ensure this fact.

**Proposition 4** *Let $P_{\mathbf{\Lambda}}(\sigma, \theta)$ be defined as in (8) and $S_j^* = L_\sigma(\nu R_j^0 + R_j)$. Then,*

$$P_{\mathbf{\Lambda}}(\theta_j \,|\, S_j^*) = Beta_{S_j^*, \nu+1}(e^{-\theta_j}) \quad (9)$$

$$P_{\mathbf{\Lambda}}(\sigma) \propto \prod_{j=1}^t Beta(S_j^*(\sigma), 1 + \nu) \quad (10)$$

*where $Beta_{\alpha,\beta}$ denotes the Beta distribution and $Beta(x, y)$ denotes Euler's Beta function.*

**Proof sketch** Replacing $\psi(\theta_j)$ with its value yields

$$P_{\mathbf{\Lambda}}(\theta_j \,|\, \sigma) \propto e^{-S_j^* \theta_j}(1 - e^{-\theta_j})^{N+\nu} \quad (11)$$

From which the desired results follow. $\square$

We have shown thus that closed form integration over the continuous parameters $\theta_j$ is possible. This result is entirely new, as no analog result, and no closed form integration is possible for the GM model with $n$ finite.

The exact summation over the discrete parameters poses much harder challenges, described in the full paper [10], and is still an open problem.

## 7 Experiments

### 7.1 Estimation experiments, single $\theta$

In these experiments we generated data from an infinite GM model with constant $\theta_j = \ln 2, \ln 4$ and estimated the central permutation and the parameter $\theta$. To illustrate the influence of $t$, $t_\pi$ was constant over each data set. The results are summarized in Table 1.

Note the apparent lack of convergence of the $\sigma^{ML}$ estimates. This is due to the fact that, as either $N$ or $t$ increase, $n_{items}$ the number of items ranked increases. The least frequent items will have very little support from the data and will be those misranked. We have confirmed this by computing the distance between the true $\sigma^{-1}$ and our estimate, restricted to the first $t$ ranks. This was always 0, with the exception of $\bar{\bar{n}} = 200, \theta = 0.69, t = 2$ when it averaged 0.04 (2 cases in 50 runs).

Even so the table shows that most ordering errors are no larger than 1. We also note that the sufficient statistic $R$ is an unbiased estimate of the expected $R$. Hence, for any fixed length $\tilde{t}$ of $\sigma^{ML}$, the $\sigma^{-1}$ estimated from $R$ should converge to the true $\sigma^{-1}$ (see also [7]). The $\theta^{ML}$ based on the true $\sigma^{-1}$ is also unbiased and asymptotically normal. Another peculiarity is the "asymmetry" of the error in $\theta^{ML}$. Recall that by equation (6) $\theta$ is a decreasing function of $L_\sigma(R)$. If the true $\sigma^{-1}$ is not optimal for the given $R$, due to sample variance, then $\theta^{ML}$ will tend to overestimate $\theta$. Hence $\theta^{ML}$ is a *biased* estimate of $\theta$. If however, due to imperfect optimization, the estimated $(\sigma^{-1})^{ML}$ is not optimal and has higher cost than $\sigma^{-1}$, then $\theta^{ML}$ will err towards underestimation.

### 7.2 Estimation experiments, general $\theta$

We now generated data from an Infinite GM model with $\theta_1 = \ln 2$ or $\ln 4$ and $\theta_j = 2^{-(j-1)/2}\theta_1$ for $j > 1$. As before, $t_\pi$ was fixed in each experiment at the values 2, 4, 8. As the estimation algorithm has local optima, we initialized the $\theta$ parameters multiple times. However, we observed that in all the experiments on artificial data, the iterations converged to the same parameter values for all initializations.

Figure 4 shows the estimated values of $\theta_j$ for sample sizes $N = 200$ and $2000$ and for the case $\theta_1 = 0.69$ (the more dispersed distribution) and $t = 8$. The results of the other experiments are similar, and they are presented in the full paper.

Qualitatively, the results are similar to those for single $\theta$, with the main difference stemming from the fact that, with decreasing $\theta_j$ values, the sampling distribution of the data is much more spread, especially w.r.t the lower ranks. Therefore, the bias in $\theta_j$ is more pronounced for larger $j$ (and for smaller $N$ and smaller $t$). For the same reason, the number of observed items $n$ is much larger than before (hundreds vs. less than 20) and consequently the ordering errors w.r.t all the observed items are also much larger[4]. But, if one considers only the top-$t$ part of $(\sigma^{-1})^{ML}$ then the results are as good as those for a single $\theta$, i.e 2 errors in 50 runs for $t = 2, \theta = 0.69, N = 200$ and *zero errors in all other trials*. We have experimented with $t$ up to 22, estimating 22 $\theta_j$ parameters, with similar results.

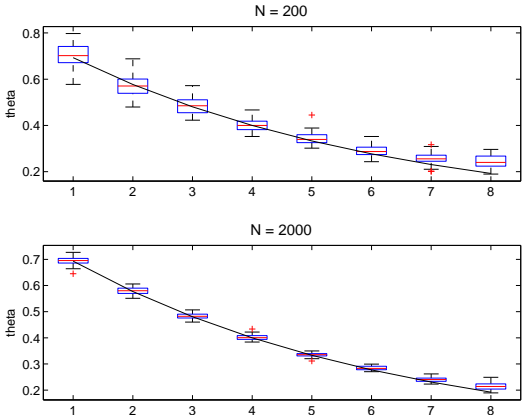

Figure 4: Example of $\theta_{1:t}$ parameters estimation; $t = 8$, sample sizes $N = 200, 2000$. Boxplots over 50 random samples, with $j$ on the horizontal axis. The continuous line crossing the box plots marks the true values of the parameters $\theta_{1:8}$ (exponential decay starting from $\theta_1 = 0.69$).

The next experiment was conducted with the data collected by Cohen, Schapire and Singer for their [3] paper. The data consists of a list of 157 universities, the queries, and a set of 21 search engines, the "experts". Each search engine outputs a list of up to $t_{max} = 30$ URL's when queried with the name of the university. The data set provides also a "target" output for each query, which is the university's home page.

Hence, we have 147 estimation problems (10 universities with no data), with sample size $N \leq 21$ (as some experts return empty lists) and with variable length data ranging from $t = 1$ to $t = 30$. Figure 5, a and b give a summary view of number of samples for each rank $N_j$, $j = 1 : t$, and respectively the total number of items $n$ per query. These values suggest that estimating a fully parameterized model with distinct $\theta_{1:30}$ may lead to overfitting and therefore we estimate several parameterizations, all having the form $\Theta_r = (\theta_1, \theta_2, \ldots \theta_{r-1}, \theta_r, \theta_r, \ldots \theta_r)$. In other words, ranks $1 : r - 1$ have distinct parameters, while the remaining ranks share 1 parameter $\theta_r$. We call $\theta_{1:r-1}$ the *free* parameters and $\theta_r$ the *tied* parameter. For

[4]Complete results are in the full paper [10].

$r = 1$ we have the single parameter model, and for $r = t_{max} = 30$ we have the fully parameterized model.

Estimating a model with $r$ parameters is done by a simple modification of the ESTIMATESIGMATHETA algorithm which is left to the reader. The estimation algorithm was started from the fixed value $\theta_j = 0.1$ for all runs. The number of iterations to convergence was typically in the range 10–30.

In figure 5 we give a synopsis of the values of the $\theta$ parameters under different models. The single parameter models yields $\theta$ values in the range [0.007, 0.104] with the 10%, 50% and 90% quantiles being respectively 0.009, 0.018, and 0.032. The parameters $\theta$ are on average decreasing in all models, with the free parameters higher than the tied parameters for the remaining ranks. Notice also that for the models with fewer parameters the values of the free parameters tend to be higher than the corresponding values in models with more parameters. Compare for instance the values of $\theta_1$ in the two-parameter model with $\theta_1$ in the 30 parameter model.

For each query and each model size, we computed the rank of the true university home page, i.e the *target*, under the estimated central permutation $\sigma^{ML}$. Assuming the search engines are reasonably good, this rank is an indirect indicator of the goodness of a model. In addition, for each query, we selected one model by BIC and calculated the target ranks for these models. The BIC selects predominantly the single parameter model (124 out of 147 cases). Table 2 gives the results.

### 7.3 Clustering experiments

We sampled orderings from 3 clusters, each an Infinite GM model with a single spread parameters $\theta$, equal to 1.5, 1.0, 0.7 respectively. The cluster centers are random permutations of infinite many objects. A data set contains 150 samples from each cluster, plus 50 outliers. All top-$t$ ordering had the same length $t$. We experimented with $t_\pi = 4, 6, 8$.

We ran the Exponential Blurring Mean-Shift, K-means, and EM Model-based clustering algorithms 10 times on samples from this distribution. For EBMS, the scale parameter was estimated based on the average of pairwise distances.

For the K-means and model-based algorithms, we experiment with different numbers of clusters, and report the best classification error with respect to the true clustering. This puts these two algorithms at an advantage w.r.t EBMS, but as the results table 3 shows, even so the nonparametric algorithm achieves the best performance.

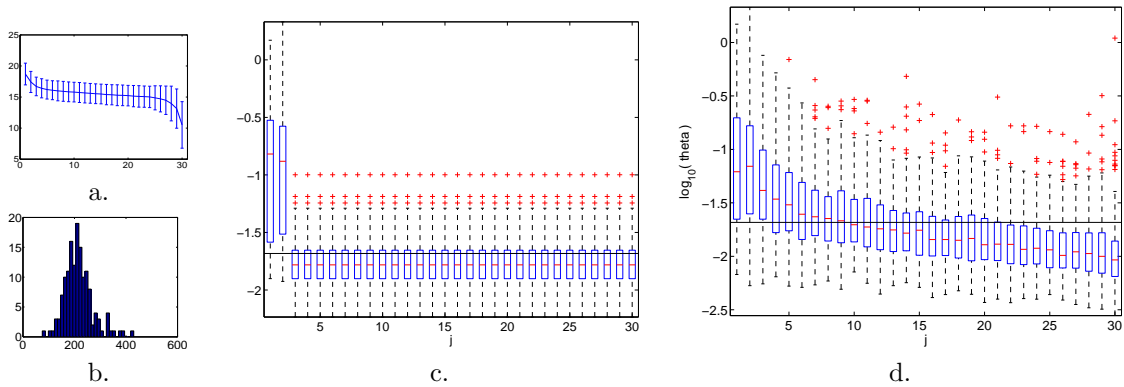

Figure 5: Universities data: mean and standard deviation of the number of samples per rank, over all queries (a); histogram of the number of distinct items observed (b); boxplots of the $\Theta$ estimates over all queries for models with 3 and 30 parameters (c,d). The vertical axis of the scale is *logarithmic, base 10*, i.e 0 corresponds to $\theta_j = 1$ and $-2$ to $\theta_j = 0.01$. For clarity, the distribution of the tied parameter is replicated for $j = r : t_{max}$. The black horizontal line marks the mean value of $\theta$ in the single parameter model.

Table 2: Mean and median of the rank of the target web page under each model, and under the BIC selected model. The rank is $t_{max} + 1 = 31$ if the target is not in the search engine's top-$t$ ranking. The statitics are computed once over all 147 universities and once over a subset of 74 universities where the target is always ranked in the first 30; the subset is labeled as "good".

| Model size | 1 | 2 | 3 | 4 | 5 | 6 | 7 | 8 | 9 | 10 | 30 | BIC |
|---|---|---|---|---|---|---|---|---|---|---|---|---|
| Mean rank (good) | 5.3 | 5.7 | 4.2 | 4.2 | **4.1** | **4.1** | 4.4 | 4.5 | 5.0 | 5.2 | 5.1 | 5.37 |
| Median rank (good) | 3 | 3 | **1.5** | 2 | 2 | 2 | 2 | 2 | 2 | 3 | 3 | 2.5 |
| Mean rank (all) | 16.5 | 16.1 | **15.4** | 15.5 | 15.5 | 15.6 | 15.8 | 15.7 | 15.9 | 16.0 | 16.0 | 16.2 |
| Median rank (all) | 13 | 15 | 11 | 11 | 12 | **9** | 10 | 10 | 11 | 11 | 11 | 12 |

Table 3: Classification Errors: mean and standard deviation of 10 random samples

| $t$ | EBMS | K-means | EM |
|---|---|---|---|
| 4 | **0.0030** (0.0001) | 0.1014 (0.0038) | 0.1008 (0.0025) |
| 6 | **0.0014** (0.0001) | 0.0986 (0.0010) | 0.1000 (0.0000) |
| 8 | **0.0002** (0.0001) | 0.0972 (0.0010) | 0.1000 (0.0000) |

Note that the error rate in table 3 is computed including the outliers, i.e we compared a true clustering with 53 clusters (3 clusters and 50 singletons) to the clustering obtained when the algorithm converged.

For EM and K-means the number of clusters associated with the lowest classification errors was between 3 and 5. From the table we see that the K-means and model based approach identified three primary clusters correctly. K-means did not have the ability of identifying the outliers, so it just assigned each outlier into one of those primary clusters. The model-based approach assigned outliers into primary clusters too, but it also gave more uncertainty on the outliers (the probabilities of outliers belonging to their assigned cluster were relatively smaller than data from primary clusters).

The running time per data set of EBMS was under a minute, and the number of iterations to convergence followed the pattern typical of mean-shift algorithms and was never larger than 10.

## 8 Related work and discussion

This work acknowledges its roots in the work of Fligner and Verducci on stagewise ordering models [6] and in the recent paper [11]. The latter shows for the first time that GM models have sufficient statistics, and describes an exact but non-polynomial algorithm to find the central permutation. While similarities exist between the algorithm of [11] and the BBOUNDR algorithm presented here, we stress that our representation (based on the codes $s_j$) is *different* from the representation (denoted $V_j$) in [11].

For any given permutation $s_j(\pi) = V_j(\pi^{-1})$. While this difference is trivial for complete permutations, it is not so in the case of missing data. In particular, the distribution of $V_j$ for top-$t$ orderings does not seem to have sufficient statistics for $j > 2$ even in the case of finite permutations. The $s_j$ representation has another advantage that $V_j$ has not: for any finite data set, a parameter $s_j$ is either completely determined or completely undetermined the data, whereas in the reciprocal $V_j$ representation *all* $V_j$ are weakly constrained by data.

While both our BBOUNDR and the algorithm of [11] perform branch-and-bound search on a matrix of sufficient statistics, the sufficient statistics in the infinite case are derived by an entirely different method, and cannot be obtained by naively replacing the sufficient statistics of the finite case.

The paper [1] uses the $s_j$ representation and outlines its advantages. It is also the first paper to do (EM) clustering of partial orderings, without however recognizing the existence of sufficient statistics. Another interesting application of the GM model to multimodal data is [9] (there the $\sigma$'s play the role of the data), while a greedy algorithm for consensus ordering with partially observed data is introduced in [3]; [4] is an early work on (Haussdorf) distances for partial orderings. In [8] the authors also introduce an EM algorithm for clustering ranking data for the purpose of analyzing Irish voting patterns. However, the base model used by [8] is not the Mallows model but a model known as Plackett-Luce. The estimation of this from data is much more difficult and, as [8] show, can be only done approximately.

All the above works deal with permutations on finite sets. In fairness to [3] we remark that this work, although it only considers heuristics methods for optimization and introduces a cost function which only later, by [11] is shown to be closely related to the log-likelihood, is motivated by the same problem as ours, i.e dealing with a very large set of items, of which only some are ranked by the "voters".

The paper of [13] studies the space of infinite permutations which differ from the identity in a finite number of positions. In the vocabulary of the present paper, these would be the infinite permutations at finite distance $d_K$ from $\sigma$. In a single parameter infinite GM, these infinite permutations are the only ones which have non-zero probability. While from a probabilistic perspective the two views are equivalent, from a practical perspective they are not. We prefer to consider in our sample space all possibile orderings, including those with vanishing probability. It is the latter who are more representative of real experiments. For instance, in the university web sites ranking experiment, our model assumed that there is a "true" central permutation from which the observations were generated as random perturbations. This is already an idealization. But we also have the liberty to assume that the observations are very long orderings which are close to the central permutation only in their highest ranks, and which can diverge arbitrarily far from it in the latter ranks. We consider this a more faithful scenario than assuming in addition that the observations must be identical to the central permutation (and hence to each other!) on all but a finite number of ranks.

We have introduced the first –to our knowledge– stage-wise ranking model for infinitely many items. The new probabilistic model has several attractive properties: it handles naturally truncated top-$t$ orderings, it has sufficient statistics, and more importantly we showed that it also has an exact estimation algorithm (albeit intractable in the worst case). As it is known from the study of stochastic models of permutations over finite domains, exact estimation and interpretable parameters are very rare qualities in this field.

Sampling, distance computations, clustering can be performed in this class of models in a natural way and are all tractable. We have paid particular attention to non-parametric clustering by mean-shift blurring, showing by experiments that the algorithm is practical and effective.

An issue not solved for GM models, finite or infinite, is sampling a $\theta, \sigma$ from the conjugate distribution. If this is feasible, one can perform clustering by the DP mixture model, a model-based clustering paradigm widely recognized for its advantages. It is our intention to work in this direction.

**Acknowledgments**

Thanks to Jon Wellner for bringing up the question of infinite permutations.